# Mimicry is Presidential: Linguistic Style Matching in Presidential Debates and Improved Polling Numbers

Daniel M. Romero*

*University of Michigan*

*Northwestern University and NICO*

Roderick I. Swaab*

*INSEAD*

Brian Uzzi

*Northwestern University and NICO*

Adam D. Galinsky

*Columbia University*

*Equal contribution

We acknowledge the funding support of the Northwestern University Institute on Complex Systems (NICO); the Army Research Laboratory under cooperative Agreement Number W911NF-09-2-0053 and Defense Advanced Research Projects Agency grant BAA-11-64, Social Media in Strategic Communication. The views and conclusions contained in this document are those of the authors and should not be interpreted as representing the official policies, either expressed or implied, of the Army Research Laboratory or the U.S. government

Word count: 5,534

**Abstract**

The current research used the contexts of U.S. Presidential debates and negotiations to examine whether matching the linguistic style of an opponent in a two-party exchange affects the reactions of third-party observers. Building off communication accommodation theory (CAT), interaction alignment theory (IAT), and processing fluency, we propose that LSM will improve subsequent third-party evaluations because matching an opponent's linguistic style reflects greater perspective taking and will make one's arguments easier to process. In contrast, research on status inferences predicts that language style matching (LSM) will negatively impact third-party evaluations because LSM implies followership. We conduct two studies to test these competing hypotheses. Study 1 analyzed transcripts of US presidential debates between 1976 and 2012 and found that candidates who matched their opponent's linguistic style increased their standing in the polls. Study 2 demonstrated a causal relationship between LSM and third party observer evaluations using negotiation transcripts.

Every Presidential candidate faces the same challenge in their high-stakes televised debates: Is it better to chart one's own linguistic path or to match the style of one's partner? This question is especially highlighted in mixed-motive situations like debates or negotiations where each person is trying to gain a competitive advantage. Past research has found that matching the content of other's language can powerfully shape the outcomes of dyadic exchanges. For example, negotiators secure better outcomes when they linguistically match their opponent (Swaab, Maddux, & Sinaceur, 2011). Opponents see negotiators who match their content as more trustworthy and it is this increased trust and likeability that leads them to make greater concessions (Miller, 2007).

A related question concerns the consequences of language style matching (LSM) on third-party observers. LSM refers to "the degree to which two people in a conversation subtly match each other's speaking or writing style" (Ireland, Slatcher, Eastwick, Scissors, Finkel, & Pennebaker, 2011, p.39) and has been found to increase group cohesion and performance (Gonzales, Hancock, & Pennebaker, 2010), success in resolving police hostage negotiations (Taylor & Thomas, 2008), and relationship initiation and stability (Ireland et al., 2011).

Although LSM research reveals how LSM differences *between* dyads affect dyadic and group level outcomes (e.g. relationship stability, Ireland & Pennebaker, 2010), it is unclear how LSM differences *within* dyads influence the evaluations of third-party observers. This class of human behavior is large and consequential. In democracies, voters evaluate candidates during debates to determine who they think is most fit to lead their nation. Where there is the rule of law, justice is rendered by juries after they assess the interactions between counsels, judges, and witnesses. Leaders often base their decisions on the negotiations between their advisors or contacts in their network (Saavedra, Duch, and Uzzi 2011).

The purpose of the current research is to investigate how LSM differences within an interaction predict the evaluations of third-party observers. Building off communication accommodation theory (CAT), interaction alignment theory (IAT), and processing fluency, we propose that LSM will improve subsequent third-party evaluations because matching an opponent's linguistic style reflects greater perspective taking and the matching will make one's arguments easier to process. In contrast, research on status inferences predicts that language style matching (LSM) will negatively impact third-party evaluations because LSM implies followership. We use the contexts of Presidential debates and job negotiations to test two competing set of hypotheses.

**Effects of Language Style Matching on Third Party Observers**

Based in the communication literature, we predict that matching the style of an opponent's language in a debate or negotiation reflects greater perspective taking and will positively affect the evaluations of third-party observers. To make this hypothesis, we draw on a set of communication theories, which propose that people coordinate their language use when engaged in a conversation. Communication Accommodation Theory (CAT) predicts that people strategically negotiate the social distance between themselves and their communication partners by matching speech patterns (convergence) or accentuating linguistic differences (divergence) (Giles & Coupland, 1991). According to CAT, linguistic convergence like LSM signals greater engagement and facilitates language processing and understanding (Coupland and Giles, 1988). Likewise, Interaction Alignment Theory (IAT) suggests that people have an innate tendency to align their word choices during their conversation, and that such alignment will promote a shared understanding of the issues at hand (Garrod & Pickering, 2004).

To capture the process of accommodation and alignment, researchers have developed a measure of Language Style Matching (LSM). The foundation of the LSM measure is the observation that predictive elements of language are words that capture the style rather than content of an utterance (Pennebaker, 2011). Whereas content related words (e.g., nouns, regular verbs) convey 'what' people say, style related words—also known as function words (e.g. prepositions and pronouns) reflect 'how' something is said (Groom & Pennebaker, 2002). Function words are therefore inherently social and require social knowledge to understand and use (Meyer & Bock, 1999; Ireland & Pennebaker, 2010): when two speakers converge in their function word choices, they are likely to share a common understanding and conceptualization of their conversation topics (Pennebaker, 2011). LSM captures this convergence by measuring the degree to which two people match function words. Further, it is important to note that high LSM levels between two speakers do not reflect rapport and cooperation per se. Although some studies have shown that greater LSM is associated with more cooperative behaviors (e.g. Gonzales et al., 2010), people locked in a bitter dispute tend to talk in similarly angry ways (Brett, Olekalns, Friedman, Goates, Anderson, & Lisco, 2007), which can result in high LSM levels as well (Ireland & Pennebaker, 2010).

CAT/IAT suggests that matchers will be perceived as more influential than non-matchers by third-party observers because LSM demands social knowledge and skill to use. That is, when one speaker tries to influence their opponent, they may work harder to read, understand, and thus better coordinate with the opponent through greater linguistic matching (Hancock, Curry, Goorha, & Woodworth, 2008). As a result, greater LSM may result in more positive evaluations because it signals that the matcher takes the opponent's perspective and is therefore in a better position to be persuasive. This reasoning is consistent with findings showing that students who matched the language their targets more (i.e. teachers) performed better (i.e. earned higher grades) (Ireland & Pennebaker, 2010) and that perspective taking increases a speaker's ability to discover opponents' preferences and to both create and claim resources in negotiations (Galinsky, Maddux, Gu, & White, 2008).

In addition to greater perspective taking, LSM may also increase third-party evaluations because matching facilitates the ease of processing and makes the content of matchers' responses seem more appealing (Day & Gentner, 2007). Indeed, language that can be processed more fluently is rated as more truthful, accurate, and persuasive than non-fluent stimuli (Alter & Oppenheimer, 2009). In turn, processing fluency, the subjective ease with which people process information (Alter & Oppenheimer, 2009), strongly affects judgment across a wide range of studies. For example, previously seen words are judged to be better answers to trivia questions (Kelley & Lindsay, 1993) and trivia statements that were repeated were judged as truer than non-repeated statements (Hasher, Goldstein, & Toppino, 1977). Recognition and repetition breed familiarity and fluency that produces "illusory truth" perceptions (Begg, Anas, & Farinacci, 1992; Whittlesea, 1993). Processing fluency also explains why rhymed phrases are more likely to be remembered than non-rhymed phrases because they are easier to process (McGlone & Tofighbakhsh, 2000). In addition, this fluency may make the matcher appear more in tune with their opponent, further raising third party evaluations.

Whereas CAT/IAT and the fluency literature both suggest that speakers who match more would be perceived as more effective, an alternative hypothesis, based in the status-inferences literature, would predict that third-party evaluators would disparage linguistic style matchers. Because the matcher follows the linguistic constructions of the other person, rather than using their own constructions, matching may be viewed as expressing deference rather than leadership.

For example, a study of the Larry King talk show found that Larry only accommodated and matched the vocal characteristics of high-status guests whereas lower-status guests accommodated to Larry (Gregory & Webster, 1996). Similar LSM associations have been found for attorneys pleading cases before Supreme Court justices and Wikipedia administrators interacting with non-administrators (Danescu-Niculescu-Mizil, Lee, Pang, & Kleinberg, 2012). These findings suggest that observers may unfavorably judge persons who match the linguistic style of their rivals during a debate or negotiation. For example, in a Presidential debate, where the candidates are trying to express authority, likely voters may disapprove of those candidates who match the linguistic style of their opponent because matching suggests the candidate is a follower, deferring to the true leader in the debate.

Although the status inferences literature predicts that matching may signal deference and thereby hurt the matcher in the eyes of their-party observers, the CAT/IAT and the processing fluency literatures predict that matchers will be perceived as more effective. In the context of third parties evaluating the competing expressions of others, these theories predict that LSM signals that matchers are better perspective takers and make their arguments more fluent and easy to understand.

## Overview

The current research allows us to test the competing hypotheses that the status inferences literature and CAT/IAT and processing fluency literatures make regarding the effect of LSM in a mixed-motive interaction on third-party evaluations. We tested these hypotheses across two studies that used both archival and experimental methods.

These studies add an important contribution to the literature because prior LSM research has only investigated how third parties evaluate the quality of the interaction process (Ireland &

Pennebaker, 2010; Niederhoffer & Pennebaker, 2002). The current research, in contrast, demonstrates the effect of LSM on the perceived effectiveness of the interaction members.

Study 1, an archival study, used all transcribed US presidential debates between 1976 and 2012 to examine the impact of LSM on third party evaluations (i.e. subsequent poll ratings). Study 2, an experimental study, manipulated LSM in a simulated job negotiation and examined its impact on third party evaluations.

## Study 1: LSM in Presidential Debates

To examine whether LSM influences the evaluations of third-party observers of an exchange, we collected all the presidential-debate transcripts that were available from 1976 to 2012 (http://www.debates.org/index.php?page=debate-history). Prior to 1976, there were a few debates and they were unsystematically conducted. Presidential debates furnish methodological advantages for our study: Specific debate questions are unknown until asked, order of exchanges is randomly assigned to candidates, and there is a quantifiable outcome: the change in polls before and after the debate. Substantively, presidential debates are increasingly studied worldwide for their impact on democracy and collective behavior (Geer, 1988; Holbrook, 1999). From 1976 to 2012, there were a total of 26 debates and 17 debaters.[1] To measure the public's favorable or unfavorable reaction to a debater, we used Gallup poll data, which is a random sample of registered voters taken at various times over the course of the presidential race, including times prior to and after the debates (http://www.gallup.com).

### Measuring LSM

We measured the extent to which candidates matched the linguistic style of the other participants in the debate (i.e., opponents, moderators, and questioners) when they are interacting with each other. We adopted a similar measure and set of procedures to the one used in (Danescu-Niculescu-Mizil, Lee, Pang, & Kleinberg, 2012), where they measured LSM on a set of 8 different linguistic markers (M) known as function words: quantifiers, conjunctions, adverbs, auxiliary verbs, prepositions, articles, personal pronouns, and impersonal pronouns (See Table 1). These linguistic markers are often chosen to measure LSM because they have little lexical meaning; hence they measure linguistic style in speech in a context- and content-free manner (Pickering & Garrod, 2004; Gonzales, Hancock, & Pennebaker, 2010). The Linguistic Inquiry Word Count (LIWC) content analysis dictionary, a validated English word classification instrument for different word categories, was used to categorize each word (Pennebaker, Booth, & Francis, 2007). Given an utterance *u* by person *p*, we say that *u* contains marker *m* or that *p* used marker *m* if *u* contains any of the words that belong to marker *m*. We give a matching score to each candidate with respect to each marker. For each debate *d*, candidate *c*, and marker *m*, we approximate the conditional probability $P_m(c,d)$ that *c* uses *m* after the person who spoke right before him used *m*. To approximate $P_m(c,d)$, we let $prev(c,d,m)$ be the number of times the person to speak right before *c* used *m*, and we let $\overline{prev(c,d,m)}$ be the number of times *c* used *m* and the person who spoke right before also used *m*. The probability $P_m(c,d)$ can therefore be approximated by $P_m(c,d) = \frac{\overline{prev}(c,d,m)}{prev(c,d,m)}$ when $prev(c,d,m) > 0$. This procedure segments the full length of the debate into sub-segments where there is a lead statement -- a question or a statement by a participant -- that is then rated for linguistic markers, and then the response statement by the next candidate is categorized as containing the marker or not. This is done for all sub-segments of a debate.

The conditional probability $P_m(c,d)$ depends on the personal style of *c* and on the overall frequency in which candidates use marker *m*. That is, $P_m(c,d)$ depends on the number of utterances by *c* that contain *m* regardless of whether the previous utterance contains *m*. For example, if all of *c*'s utterances contain *m* then $P_m(c,d) = 1$. On the other hand, if none of *c*'s utterances contain *m* then $P_m(c,d) = 0$. To insure that the LSM in any exchange was not due to chance use of words, we used coincidence analysis, which takes the observed utterances of a debater, randomizes them, and then calculates a z-score for the observed and randomized utterances to see how far from chance the observed utterance was. Specifically, we compared $P_m(c,d)$ with its corresponding value when the ordering of the debater's utterances is randomized. If *c*'s usage of *m* tends to match to the utterances spoken right before his, then $P_m(c,d)$ should be significantly different when the utterances are in their actual order than when they are in random order. We let $D_m(c,d)$ be the collection of 10,000 $P_m(c,d)$ values, each one corresponding to one of 10,000 random orderings of the debate utterances. We compare $P_m(c,d)$ with $D_m(c,d)$ by computing the z-score, $z_m(c,d) = \frac{mean(D_m(c,d)) - P_m(c,d))}{std(D_m(c,d))}$. The value of $z_m(c,d)$ indicates the extent to which candidate *c* was matching to others during debate *d*. The larger $| z_m(c,d) |$ is, the higher our confidence that *c* was matching if $z_m(c,d) > 0$, or mismatching, if $z_m(c,d) > 0$ during the debate *d*. Finally, we measure the central tendency of linguistic matching of *c* during debate *d* by taking the mean of $z_m(c,d)$ for the 8 different markers *M*. We denote the mean of the 8 z-scores as $z(c,d) = \frac{1}{M} \sum_{m \in M} z_m(c,d)$.

In summary, the following steps are used to compute LSM for debate *d*, candidate *c*, and marker *m*:

1. Compute $P_m(c,d)$, the fraction of times *c* used marker *m* over the number of times the person who spoke right before *c* used *m*.

2. Randomize utterances in debate 10,000 times
3. For each randomization, compute the corresponding value of $P_m(c,d)$.
4. Let $D_m(c,d)$ be the collection of 10,000 $P_m(c,d)$ values.
5. The measure of LSM for marker m is the z-score

$$z_m(c,d) = \frac{mean(D_m(c,d)) - P_m(c,d))}{std(D_m(c,d))}.$$

The measure of LSM we used gives a score to each candidate that captures the extent to which the candidate changed her personal style to match the style of her opponents by computing the probability of word use, regardless of the number of times the word is used within each sentence. In contrast, other measures of LSM, such as the one used by Ireland and Pennebaker, compute the similarity in the percentage of words used by two speakers. The percentage of word use is an appropriate measure for exchanges where each statement is very long such as exchanges of letters (Ireland & Pennebaker, 2010). However, for exchanges of short statement such as debates, percentage of word use is very often zero, and thus inappropriate. We computed Ireland and Pennebaker's turn-by-turn LSM measure (Ireland & Pennebaker, 2010) and found that it is positively correlated with our measure (r = 0.62).

| Word Category | Size | Examples |
|---|---|---|
| Quantifiers | 20 | all, remaining, somewhat |
| Conjunctions | 28 | also, but, unless |
| Adverbs | 68 | about, especially, perhaps |
| Auxiliary verbs | 147 | am, must, might |
| Prepositions | 60 | about, besides, near |
| Articles | 4 | a, an, the |
| Personal pron. | 71 | he, she, our |
| Impersonal pron. | 46 | anybody, these, it |

**Table 1.** Number of words in each category and examples.

**Third-Party Evaluations**

To investigate the relationship between LSM and third party reactions, we used the results from the Gallup presidential-race polls. These polls were conducted on various dates starting a few months before the election until the day of the election. We measured the effect of LSM on the polls by comparing poll results before and after each debate. Since any individual poll gives a noisy signal of the popularity of the candidates and we do not have access to the margin of error of the polls, we do not base our measure on simply the difference between the polls immediately before and after each debate. Instead, we take the difference between the median result among multiple polls taken before and after each debate. This provides a more robust signal of how the popularity of the candidates changed after the debate. To account for trends and autocorrelation bias in a candidate's poll numbers, we measured changes as difference scores (Granger, 1969). More precisely, for each race with *n* debates $d_1 \ldots d_n$, which occurred on dates $t_1 \ldots t_n$, we let $t_0$ be September 1st and $t_{n+1}$ be the day of the election. For each debate $d_i$ and candidate *c*, we let $P_b(d_i, c)$ and $P_a(d_i, c)$ be the median poll results for candidate *c* during the time period $(t_{i-1}, t_i)$ and $(t_i, t_{i+1})$, respectively. The quantity $P_{diff}(d_i, c) = P_a(d_i, c) - P_b(d_i, c)$ measures how the polls changed from before to after debate *d* after accounting for trends in the polls.

**Results**

Figure 1 shows the bivariate relationships between linguistic matching, non-matching and change in polls $P_{diff}$ for the debate. The scatter plot shows that increases in LSM are consistently and positively related to that candidate's subsequent increase in the polls.

Focusing on the effect size of LSM, we compared the poll changes in cases when candidates displayed LSM during the debates with those cases when they displayed no matching.

We defined a set of *matchers* $M = \{(c,d) : z(c,d) > 0\}$ as cases when a candidate had a positive mean of LSM z-scores, and a set of *non-matchers* $\overline{M} = \{(c,d) : z(c,d) < 0\}$ as cases when a candidate had a negative mean of LSM z-scores.

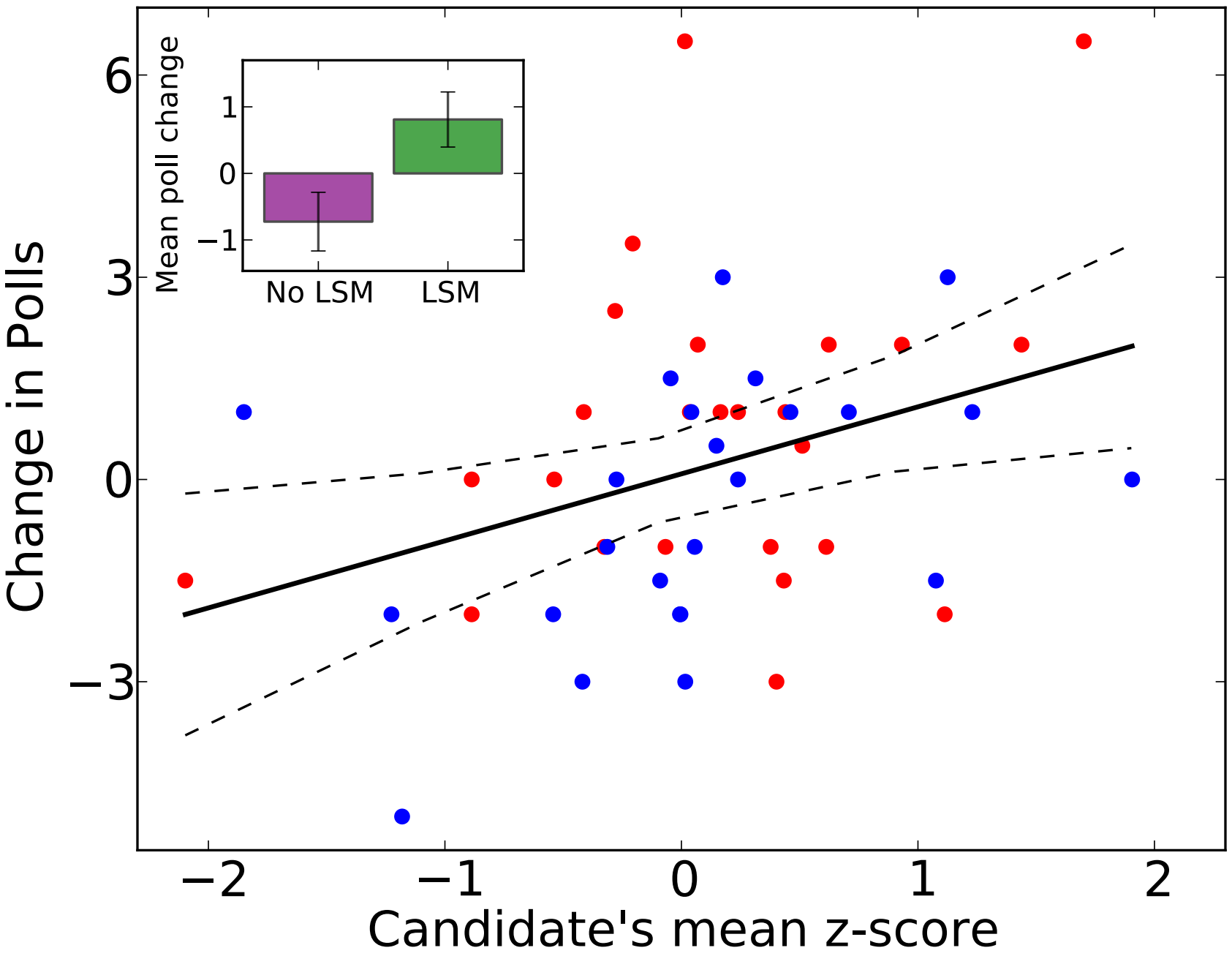

**Figure 1. Candidate's mean LSM z-scores and Change in Polls.** Paired values for $(z(c,d), P_{diff}(c,d))$ with a simple linear regression and 95% confidence interval. Red and blue dots represent Republican and Democratic candidates, respectively. The subplot shows the average change in polls for linguistic matchers and non-matchers split at a z-score of 0.0 with 95% confidence intervals. The difference between linguistic matching and non-matching is significant (p-val < 0.01).

Figure 1 inset shows the average change in polls $P_{diff}$ for matchers and non-matchers. We find that the median gain for matchers is 1 point and the median loss for non-matchers is 1 point (Mann-Whitney U test for difference in medians, $p$=0.017) while the simple mean gain for matchers is 0.81 points and simple mean losses for non-matchers is 0.73 points in the poll numbers (t-test for difference in means, $p$ = 0.016). This suggests that linguistic matching appears to gain favorable impressions from 3rd party observers and vice versa for linguistic mismatching.

Changes in polls may be affected by heterogeneity in candidates' characteristics or election year characteristics. To conservatively control for this heterogeneity, we used fixed effects regression models (Laird & Ware,1982; McCaffrey, Lockwood, Mihaly, & Sass, 2012), which estimate the net effect of LSM on poll changes after accounting for other person-level and election year level factors affecting poll change; i.e., the model estimates within rather than between person effects of LSM on 3rd party impressions. For example, individual fixed effects control for all person level characteristics that are unobserved and unchanging such as charisma, physical characteristics like good looks, height, IQ, and pitch of voice, or habituated mannerisms such as twitching and eye blinking such that only the residual variance in poll change can be explained by LSM (Ballew & Todorov, 2007). Election year fixed effects control for the state of economy, wartime, and so forth (Healy, Malhotra, & Mo, 2012).

Given $n$ individuals or units and $T$ observations coming from each unit, a fixed-effects regression model takes the form $y_{it} = X_{it}\beta + \alpha_i + e_{it}$ for $t = 1,...,T$ and $i = 1,...,n$, where $y_{it}$ is the dependent variable for observation $t$ coming from unit $i$, $X_{it}$ is the regressor matrix, $e_{it}$ is the error term, and $\alpha_i$ is the unobserved unit-invariant effect for unit $i$. The model attributes co-variance between cases and $y_{it}$ to the term $\alpha_i$, instead of attributing it to the $\beta$ of an independent variable as the simple linear regression model would.

The fixed effects regression tests provided strong evidence that linguistic style matching is significantly and positively related to favorable third party reactions net of other fixed factors known to affect poll changes. Consistent with our bivariate results, LSM had a significant and positive effect on the subsequent change in polls. The adjusted $R^2$ for the most conservative regression containing both candidate and election fixed effects was 0.55 and the LSM coefficient was 0.76 ($p$ = .019). This suggests that LSM is viewed positively by third parties to a debate. Whether one examines the bivariate relationship at the mean or medians or with fixed effects for persons and elections, the results indicate that linguistic matching results in favorable audience responses.

**Discussion**

Study 1 found that greater LSM during presidential debates predicted favorable poll movement. These findings support the predictions based in the CAT and processing fluency literatures. While the effect of LSM in Presidential debates is important in and of itself, the processing advantage of fluent information in this context could be correlated with other factors that we were not able to control for statistically. For example, processing advantages of fluent information increases with age (Skurnik, Yoon, Park, & Schwarz, 2005). Because older citizens are more likely to watch debates than younger voters (Kenski & Stroud, 2005; Kenski & Jamieson, 2008), the fluency benefits of linguistically matching one's opponent may be particular to the U.S. Presidential debate context. To test whether the findings in Study 1 are specific to political debates, we conducted an experimental study in the context of job negotiations.

## Study 2: Causal Effect of LSM on Third-Party Evaluations

To test whether LSM has a causal impact on third party evaluations, Study 2 manipulated LSM in the context of a negotiation. We examined whether greater LSM would predict how positively negotiators are evaluated by third-party observers.

**Participants, Design, and Method**

Seventy-nine participants (24 males, 55 females; mean age=40.11 years, SD=17.48) were recruited from a U.S. University online pool and randomly assigned to read a negotiation transcript where the candidate mimicked more than the recruiter (candidate-LSM condition) or the recruiter mimicked more than the candidate (recruiter-LSM condition). Nine participants did not respond correctly to an attention check and were omitted from the analyses. Including them does not affect the results reported below. Participants were not given a time limit for reading their assigned transcript, but could not proceed with the survey until after 4 minutes had passed.

**LSM Manipulation**

We conducted a pre-test to create the LSM manipulations and test whether greater LSM would yield more favourable third party evaluations. We instructed 88 MBA students, enrolled in a negotiations course at a global business school, to participate in a text-based, online, simulated job negotiation between a recruiter and a job candidate ("New Recruit", Neale, 1997). Negotiations were conducted in an online, text-based format to remove any impact of body language, gender, age, or attractiveness of negotiators. Thus, the *only* way in which negotiators could mimic each other was through the exchange of words.

LSM scores were calculated in the exact same way as in Study 1. We selected two transcripts from this initial sample to create the LSM conditions, one in which the recruiter mimicked more than the candidate and a second one in which the candidate mimicked more than the recruiter. We used four criteria to select these two transcripts for our experiment: 1) candidates (recruiters) linguistically mimicked the recruiter (candidate) significantly more than the other way around, 2) there were no differences in outcomes between the negotiators (i.e. the candidate and recruiter achieved an equally profitable agreement in the negotiation), 3) there were no qualitative differences in the use of affect-based language (i.e. positive and negative emotion words) between the negotiators and 4) LSM asymmetry was equivalent between negotiators. This approach allowed us to test whether greater LSM would result in more favorable third party evaluations and establish that this would occur independent of negotiator role, negotiation outcome, communication valence, and LSM asymmetry.

Two transcripts, one in which the candidate linguistically mimicked the recruiter and one in which the recruiter linguistically mimicked the candidate, met these criteria. All identifying information of the negotiators and information about outcomes were removed from the transcripts. We conducted our experiment using these and only these transcripts.

The use of affect-based language using LIWC was similar across these two transcripts (6.82% vs. 6.04% of the total word count for transcripts 1 and 2, respectively). The percentage of emotionally positive and negative words was 6.04% and 0.8% respectively, in transcript 1, and 4.7% and 0.09% respectively, in transcript 2. Hence, the use of positive words dominated negative words in both transcripts. We also consider negation words such as "no" and "never" and assent words such as "yes" and "agree." The percentage of negations and assent words in transcript 1 was 0.7% and 1.2%, and 0.9% and 1.1% in transcripts 2, respectively. LSM asymmetry was equally strong in both of the selected transcripts (standardized LSM scores differed at $Z = 1.05$ in the candidate-LSM condition and $Z = 1.20$ in the recruiter-LSM condition).

We also measured differences in word use between the matching and matched negotiators. Kacewicz et al. found that high-status individuals tend to use the first-person plural pronoun "we" more than the first-person singular pronouns "I" and "me" (Kacewicz, Pennebaker, Davis, Jeon, & Graesser, 2013). We compared the use of plural and singular first person pronouns as well as emotionally positive and negative, negations, and assent words between matchers and non-matchers. We did not find a consistent difference in the use of these word categories between negotiators in the two transcripts we used. The only word category that was used in significantly different frequency by matchers and non-matchers was singular first person pronoun. However, in transcript1 the matcher used it in higher frequency that the non-matcher, and the opposite was the case in transcript 2.

Thus, these two transcripts enabled us to examine how variation in LSM influenced third-party evaluations while holding negotiator outcome, the valence of their speech, and the strength of LSM constant. Due to the online, text-based nature of the negotiation and the removal of identifying information, any effects could also not be influenced by body language, gender, age, or attractiveness of the negotiator.

**Dependent Measure**

Participants evaluated the negotiators using the following three items, *"Who do you think did the better job in the negotiation?," "Who do you think won the negotiation?,"* and *"Whom do you pick to negotiate for you?"* on a 5-point scale (1=candidate, 3=both equally, 5=recruiter (α=.91).

**Results**

Replicating the Presidential debate analysis, greater LSM led to more favorable impressions from third party observers: Candidates were evaluated more positively in the candidate-LSM condition (*M*=2.00; *SD*=.90) than in the recruiter-LSM condition (*M*=2.59; *SD*=1.13), $t(68)=2.59$, $p=.028$, $\eta^2=.07$; the same was necessarily true for the recruiter. Thus, LSM has a causal impact on third party evaluations.

## A Question of Timing: Early versus Late LSM and Favorable Third-Party Evaluations

One interesting question that emerges from this research is whether it is better to linguistically match one's opponent early or late in the exchange in order to improve third-party evaluations. Prior work has found that the effect of linguistic matching for a dyad's mixed-motive outcome is critical early in a negotiation because it lays the groundwork of trust between the mimicker and mimickee. Thus, negotiators who linguistically match their opponent early in a negotiation secure the necessary trust of their opponent to extract concessions from that opponent (Swaab, Maddux, & Sinaceur, 2011). In terms of outcomes within the dyad, it is better to linguistically match one's opponent earlier than later in the exchange.

However, in the context of third-party evaluations, recently presented information has greater impact than earlier presented information. Recency effects explain why candidates who perform later in serial competitions get higher scores than candidates who perform at the beginning of competitions even when the order of the candidates' presentation or performance is randomized (Bruine de Bruin , 2005; Mantonakis, Rodero, Lesschaeve, & Hastie, 2009). CAT/IAT also suggests that the positive effect of matching may be more pronounced later rather than earlier because when one speaker tries to influence their opponent, it requires time to read, understand, and thus better coordinate with the opponent through greater linguistic matching (Hancock, Curry, Goorha, & Woodworth, 2008).

Study 1 allowed us to explore the temporal dynamics of LSM and test whether linguistic matching would have a greater effect when it comes later in the debate than when it comes earlier. We split each debate into 40-time-ordered parts with each part having an equal number of utterances. We measured each candidate's LSM only taking into account the first $i^{th}$ parts. Figure 2 shows the mean $z(c,d)$ as a function of the number of parts we consider for candidates whose poll numbers go up $(P_{diff} > 0)$ and down $(P_{diff} > 0)$ separately and shows that the mean pattern of LSM matching across the debates begins with mismatching by both candidates. This figure demonstrates that candidates that have a positive change in the polls are associated with a clear and steady increase in matching over the course of the debate while candidates that drop in the polls show the opposite pattern.

These analyses reveal that candidates that matched the linguistic style of their opponents in the debate received a significant and positive change in the polls especially when the LSM occurred later in the debate.

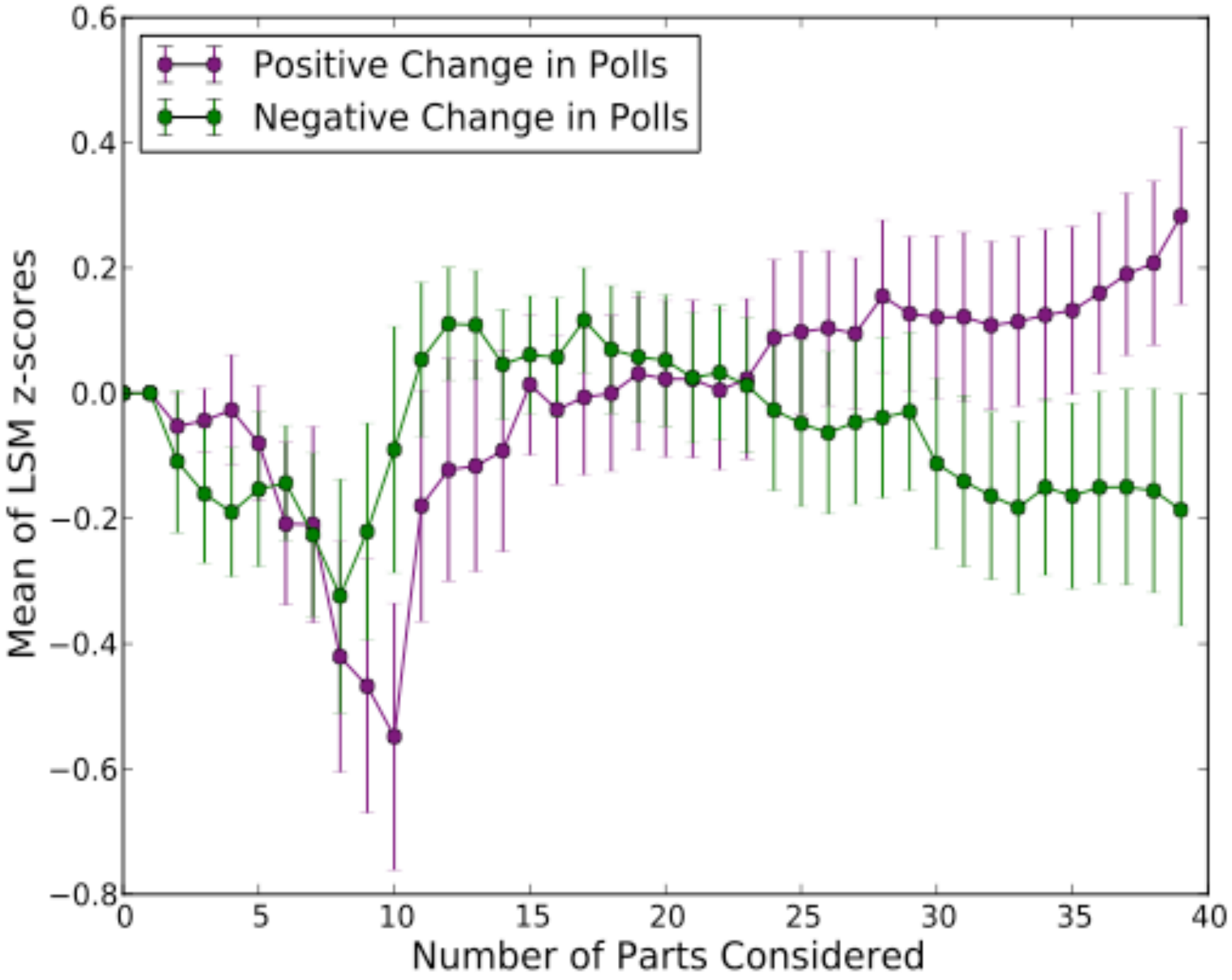

**Figure 2. Mean LSM z-scores throughout debate segments.** We split each debate into 40 time ordered parts where each part contains the same number of utterances. The figure shows the mean of linguistic matching z-scores $z$ vs. the number of consecutive debate parts considered for candidates whose poll numbers increased ( $P_{diff} > 0$ ) and decreased ( $P_{diff} < 0$ ) after the debate. The error bars correspond to the 95% confidence interval of each sample.

## Discussion

The current research explored whether linguistic style matching (LSM) would positively or negatively affect third-party evaluations in the context of Presidential debates and negotiations. Past research on linguistic matching has mostly looked at its effects within the dyad itself. For example, linguistic matchers in intimate relationships and negotiations are more liked and trusted by the other person in that exchange (Gregory & Webster, 1996; Swaab, Maddux, & Sinaceur, 2011). With regard to the impact on third-party observers, status inference theories would predict a negative effect of LSM because linguistic matching belies the candidate's authority and leadership (Danescu-Niculescu-Mizil, Lee, Pang, & Kleinberg, 2012).

In contrast, building on CAT/IAT and fluency theory, we reasoned that LSM would lead to greater approval of the matching candidate. CAT/IAT finds that greater linguistic convergence signals that matchers internalized their opponent’s thinking more and are therefore better positioned to influence them. Fluency theory has found that speakers that display greater fluency receive greater approval, less scrutiny of their verbal content, and higher levels of trustworthiness from their audience. Consistent with CAT/IAT and fluency theory, we found that higher LSM during a Presidential debate and a negotiation improved the evaluation of third-party observers relative to the mismatching speaker. These findings are consistent with other research demonstrating that information processing, rather than content, can impact collective decision making in electoral politics (Healy, Malhotra, & Mo, 2012).

These findings suggest that LSM relates to the performance of two debaters or negotiators in different directions depending on how performance is measured. The present studies show that when performance is measured by the perception of third-party observers, LSM positively relates to performance. However, in other settings, such as police interrogations, being matched relates to obtaining more favorable outcomes, such as obtaining a confession (Richardson, Taylor, Snook, Conchie, & Bennell, 2014).

Although the present research found support for CAT/IAT and fluency theory and not for status inference theory, it is possible that LSM negatively affects third-party evaluations as well. For instance, LSM may undermine third-party evaluations when the matcher follows the linguistic patterns of the other person exclusively at high levels. Future research could investigate more closely the conditions under which LSM undermines third party evaluations. Although prior work has found that content matching is critical early in a conversation because it lays the groundwork of trust essential for extracting concessions from one's opponent (Swaab, Maddux, & Sinaceur, 2011), the current research suggests that LSM may be more important later in an exchange in terms of influencing third-party evaluations. Future research could further explore how the timing of linguistic content- and style matching affects third-party evaluations.

## Conclusion

By focusing on the consequences of LSM on third-party observations, the current research offers an important departure from past LSM research, which focused predominantly on LSM differences within dyads on dyadic outcomes (Ireland et al, 2011). Specifically, the current research suggests that the effects of LSM have different effects within the dyad versus on third-parties observing the dyad. Third-party observers to an exchange were affected by LSM mechanisms that make it easier to process information and accept the statements of the matcher.

Linguistic matching has been argued to be unconscious (Niederhoffer & Pennebaker, 2002). This suggests that linguistic matchers may be oblivious to its impact. People, who do not match their opponent's linguistic style, perhaps by actively attempting to persuade the public by highlighting differences between them and the opposition, may misunderstand that mimicry is presidential.

## References

Alter A. L., & Oppenheimer D. M. (2009). Uniting the Tribes of Fluency to Form a Metacognitive Nation. *Personality and Social Psychology Review*, 13, 219-235.

Ballew C.C., & Todorov A. (2007). Predicting political elections from rapid and unreflective face judgments. *Proceedings of the National Academy of Sciences*, 104, 17948-17953.

Begg, I.M., Anas, A., & Farinacci, S. (1992). Dissociation of processes in belief: Source recollection, statement familiarity, and the illusion of truth. *Journal of Experimental Psychology: General,* 121, 446-458.

Bruine de Bruin W. (2005). Save the last dance for me: Unwanted serial position effects in jury evaluations. *Acta Psychologica*, 118, 245-260

Danescu-Niculescu-Mizil C., Lee L., Pang B., & Kleinberg J. (2012). Echoes of power: language effects and power differences in social interaction. *Proceedings of the 21st international conference on World Wide Web*, 699-708.

Day S. B., & Gentner D. (2007). Nonintentional analogical inference in text comprehension. *Memory & Cognition*, 35, 39-49.

Feddersen T. J., & Pesendorfer W. (1999). Elections, Information Aggregation, and Strategic Voting. *Proceedings of the National Academy of Sciences*, 96, 10572-74.

Fowler J. H., & Schreiber D. (2008). Biology, Politics, and the Emerging Science of Human Nature. *Science*, 322, 912-914.

Frederickson B. L., & Kahneman D. (1993). Duration neglect in retrospective evaluations of affective episodes. *Journal of Personality and Social Psychology*, 65, 45-55.

Garrod, S., & Pickering, M. J. (2004). Why is conversation so easy? *Trends in Cognitive Science*, 8, 8-11.

Geer J. G. (1988). The effects of presidential debates on the electorate's preferences for candidates. *American Politics Research*, 16, 486-501.

Giles, H. & Smith, P. (1979). Accommodation Theory: Optimal Levels of Convergence. In H. Giles & R.N. St. Clair (Eds.), *Language and Social Psychology*. Baltimore: Basil Blackwell.

Giles, H. & Coupland, N. (1991). *Language: Context and consequences.* Pacific Grove, CA: Brooks/Cole.

Gonzales A. L., Hancock J. T., & Pennebaker J. W. (2010). Language Style Matching as a Predictor of Social Dynamics in Small Groups. *Communication Research*, 37(1), 3-19.

Granger C.W.J. (1969). Investigating causal relations by econometric models and cross-spectral methods. *Econometrica*, 37, 424-438.

Gregory S. W., & Webster S. (1996). A Nonverbal Signal in Voices of Interview Partners Effectively Predicts Communication Accommodation and Social Status Perceptions. *Journal of Personality and Social Psychology*, 70, 1231-1240.

Groom, C. J., & Pennebaker, J. W. (2002). Words. *Journal of Research in Personality, 36,* 615-621.

Hancock, J. T., Curry, L., Goorha, S., & Woodworth, M. T. (2008). On lying and being lied to: A linguistic analysis of deception. *Discourse Processes, 45,* 1–23.

Hasher, L., Goldstein, D., & Toppino, T. (1977). Frequency and the conference of referential validity. *Journal of Verbal Learning and Verbal Behavior*, 16, 107-112.

Healy A. J., Malhotra N., & Mo C. H. (2012). Irrelevant events affect voters' evaluations of government performance. *Proceedings of the National Academy of Sciences*, 107, 12804-12809.

Holbrook T. M. (1999). Political Learning from Presidential Debates. *Political Behavior*, 21, 67-89.

Ireland, M.E., Slatcher, R.B., Eastwick, P.W., Scissors, L.E., Finkel, E.J., & Pennebaker, J.W. (2011). Language style matching predicts relationship initiation and stability. *Psychological Science,* 22, 39-44.

Ireland, M.E. & Pennebaker, J.W. (2010). Language style matching in writing: Synchrony in essays, correspondence, and poetry. *Journal of Personality and Social Psychology*, 99, 549-571.

Kacewicz E., Pennebaker J. W., Davis M., Jeon M., & Graesser A. C. (2013). Pronoun use reflects standings in social hierarchies. *Journal of Language and Social Psychology*. 33(2), 125-143.

Kelley C. M., & Lindsay D. S. (1993). Remembering mistaken for knowing: Ease of retrieval as a basis for confidence in answers to general knowledge questions. *Journal of Memory and Language*, 32, 1-24.

Kenski K., & Jamieson K. H. (2008). Presidential and Vice Presidential Debates in 2008: A Profile of Audience Composition. *American Behavioral Scientist*, 55, 307-324.

Kenski K., & Stroud N. J. (2005). Who Watches Presidential Debates? A Comparative Analysis of Presidential Debate Viewing in 2000 and 2004. *American Behavioral Scientist*, 49, 213-228.

Laird N. M., & Ware J. H. (1982). Random-effects models for longitudinal data. *Biometrics*, 38, 963-974.

Mantonakis A., Rodero P., Lesschaeve I., & Hastie R. (2009). Order in choice: effects of serial position on preferences. *Psychological Science*, 20, 1309-1312.

McCaffrey D.F., Lockwood J.R., Mihaly K., & Sass T.R. (2012). A review of Stata commands for fixed-effects estimation in normal linear models. *The Stata Journal*, 12, 406-432.

McDermott R., Fowler J. H., & Smirnov O. (2008). On the Evolutionary Origin of Prospect Theory Preferences. *The Journal of Politics*, 70, 335-350.

McGlone M. S., & Tofighbakhsh J. (2000). Birds of a feather flock conjointly (?): Rhyme as reason in aphorisms. *Psychological Science*, 11, 424-428.

Meyer, A. S., & Bock, K. (1999). Representations and processes in the production of pronouns: Some perspectives from Dutch. *Journal of Memory and Language*, *41,* 281-301.

Miller G. (2007). The Art of Virtual Persuasion. *Science*, 317, 1343.

Neale, M. A. (1997). New Recruit. In J. M. Brett (Ed.), *Negotiation and decision making exercises* Evanston: Dispute Resolution Research Center.

Niederhoffer K. G., & Pennebaker J. W. (2002). Linguistic Style Matching in Social Interaction. *Journal of Language and Social Psychology*, 21, 337-360.

Pennebaker J. W., Booth R. J., & Francis M. E. (2007) Linguistic inquiry and word count (LIWC): A computerized text analysis program, [Computer software]. Austin, TX: LIWC.net.

Pennebaker, J. W. (2011). *The secret life of pronouns: How our words reflect who we are*. New York, NY: Bloomsbury Press.

Pickering M. J. , & Garrod S. (2004). Toward a mechanistic psychology of dialogue. *Behavioral and Brain Sciences*, 27, 169-190.

Plackett R. L. (1983). Karl Pearson and the Chi-Squared Test. *International Statistical Review*, 51, 59-72.

Richardson B. H., Taylor P.J., Snook B., Conchie S.M., Bennell C. (2014). Language style matching and police interrogation outcomes. *Law and Human Behavior*, 38(4), 357-366.

Shaffer, J. P. (1995). Multiple Hypothesis Testing. *Annual Review of Psychology*, 46, 561-584.

Skurnik I., Yoon C., Park D. C., & Schwarz N. (2005). How Warnings about False Claims Become Recommendations. *Journal of Consumer Research*, 31, 713-724.

Saavedra S., Duch J., & Uzzi B. (2011). Tracking Traders' Understanding of the Market Using E-communication Data. *PLoS ONE 6(10)*: e26705.doi:10.1371

Swaab R. I., Maddux W. W., & Sinaceur M. (2011). Early words that work: When and how virtual linguistic mimicry facilitates negotiation outcomes. *Journal of Experimental Social Psychology*, 47, 616-621.

Taylor, P.J. & Thomas, S. (2008). Linguistic style matching and negotiation outcome. *Negotiation and Conflict Management Research,* 1, 263-281.

Turner, L.H. & West, R. (2010). Communication Accommodation Theory. *Introducing Communication Theory: Analysis and Application* (4th ed.). New York, NY: McGraw-Hill.

Tversky A., & Kahneman D. (1974). Judgment under Uncertainty: Heuristics and Biases, *Science*, 185:1124—1131.

Whittlesea, B.W.A. (1993). Illusions of familiarity. *Journal of Experimental Psychology: Learning, Memory, and Cognition,* 6, 1235-1253.

Footnotes

[1] We did not use the final debate of the 2012 election since not all the poll numbers that came after this debate were available at the time the study was conducted. Primary election debates and vice-president debates were not included in the study.